\documentclass[10pt, a4paper]{article}
\usepackage{lrec2022} 
\usepackage{multibib}
\newcites{languageresource}{Language Resources}
\usepackage{graphicx}
\usepackage{tabularx}
\usepackage{soul}
\usepackage{titlesec}
\titleformat{\section}{\normalfont\large\bfseries\center}{\thesection.}{1em}{}
\titleformat{\subsection}{\normalfont\SmallTitleFont\bfseries\raggedright}{\thesubsection.}{1em}{}
\titleformat{\subsubsection}{\normalfont\normalsize\bfseries\raggedright}{\thesubsubsection.}{1em}{}
\renewcommand\thesection{\arabic{section}}
\renewcommand\thesubsection{\thesection.\arabic{subsection}}
\renewcommand\thesubsubsection{\thesubsection.\arabic{subsubsection}}

\usepackage{epstopdf}
\usepackage[utf8]{inputenc}

\usepackage{hyperref}
\usepackage{xstring}

\usepackage{color}

\title{ \textbf{A Survey of Multilingual Models for Automatic Speech Recognition}}

\name{Hemant Yadav, Sunayana Sitaram} 

\address{Microsoft Research India \\
         Bangalore \\
         t-hyadav@microsoft.com, sunayana.sitaram@microsoft.com\\
        }

\abstract{
Although Automatic Speech Recognition (ASR) systems have achieved human-like performance for a few languages, the majority of the world's languages do not have usable systems due to the lack of large speech datasets to train these models. Cross-lingual transfer is an attractive solution to this problem, because low-resource languages can potentially benefit from higher-resource languages either through transfer learning, or being jointly trained in the same multilingual model. The problem of cross-lingual transfer has been well studied in ASR, however, recent advances in Self Supervised Learning are opening up avenues for unlabeled speech data to be used in multilingual ASR models, which can pave the way for improved performance on low-resource languages. In this paper, we survey the state of the art in multilingual ASR models that are built with cross-lingual transfer in mind. We present best practices for building multilingual models from research across diverse languages and techniques, discuss open questions and provide recommendations for future work.
 \\ \newline \Keywords{speech recognition, multilingual, low-resource languages} }

\begin{document}

\maketitleabstract







\section{Introduction}

The field of Automatic Speech Recognition (ASR) has made significant progress over the last few years due to the advent of Deep Neural Network (DNN) based models. Most notably, on benchmarks such as Switchboard, models have now achieved human parity \cite{xiong2016achieving}, which means that models are able to transcribe human speech with the same or better accuracy as humans. However, due to the large data requirement of DNN-based models, most of these advances have been restricted to a few languages of the world, which have large datasets of transcribed speech. Languages that do not have much transcribed and/or untranscribed speech, that is, low-resource languages have been left behind.

\cite{joshi2020state} classify languages based on resources for Natural Language Processing (NLP) and find that almost 90\% of the world's population speaks languages that are very low resource and are thus not benefited by language technologies. Since speech datasets are usually more expensive to build than text datasets, the situation may be even more stark for ASR. This is a lost opportunity for serving language communities where speech is the only medium of expression, owing to low literacy or the lack of a standardized writing system.

The problem of ASR for low resource languages has been well studied starting from traditional HMM-GMM based models, to the transformer-based models used today. The main technique used for addressing the lack of resources in the target language is cross-lingual transfer, in which models or resources from a higher resource language are adapted to the target low resource language using adaptation data. Cross-lingual transfer may take place using models trained on a single high resource language, or multiple languages with varying amounts of resources. The intuition behind these techniques is that the lower resource language benefits from the language invariant features learned by the model from the large amount of data in the high resource language.

Recently, there has been a paradigm shift in NLP research with the advent of pre-trained models such as BERT \cite{BERT}. These models first use unlabeled data (raw text) for pre-training and are then fine-tuned for a downstream task using labeled task-specific data. The intuition behind this approach is that the model can learn features from unlabeled data that can be shared across tasks. Massively Multilingual Language Models (MMLMs) \cite{devlin2019bert,conneau2020unsupervised} are pre-trained on a large amount of unlabeled text from multiple languages (around 100) and then fine-tuned for a particular language and task. MMLMs have shown impressive cross-lingual transfer \cite{wu2019beto}, leading to performance gains on languages that have no labeled data (through zero-shot learning) or a small amount of labeled data (few-shot learning). Recently, the idea of using unlabeled data for pre-training has also been used in the field of speech processing \cite{chung2021w2v}, which could greatly solve the challenges of low-resource ASR if successful, since there may be large amounts of untranscribed speech available in many languages. An additional benefit of multilingual models is that training and maintenance costs can be lowered by reusing the same model for multiple languages, which can benefit low-resource languages having limited budgets.

In this paper, we survey the progress made in multilingual ASR. For the purposes of this paper, we define multilingual ASR models as models that are trained with the aim of aiding cross-lingual transfer to eventually benefit other (usually low-resource) languages. We classify them as follows (i) ASR models trained using labelled data in one or multiple languages (ii) ASR models that are first pre-trained using unlabeled data from one or multiple languages and then fine-tuned using labeled data. In our survey, we set out to answer the following questions:

\begin{itemize}

    \item Are multilingual models better in terms of performance compared to monolingual models for high and low-resource languages?
    \item Can multilingual ASR models exploit unlabeled data for improving performance on low-resource languages?
    \item What are the best practices for building multilingual ASR models?
    \item Where should resources be invested for improving low-resource ASR in the era of multilingual models?
    \item What are the open questions and main challenges that need to be addressed going forward?
\end{itemize}

The paper is organized as follows. Section 2 describes the general ASR training pipeline and resources required for building ASR models. In Section 3, we survey multilingual models for ASR categorized by whether or not they use unlabeled data. In Section 4, we list the key findings and open questions that still need to be addressed. Section 5 concludes.



\section{ASR training and resources}

ASR is the task of converting a spoken utterance into a sequence of words. It can be broken down into three broad steps. 

(i) The first step is to convert the raw audio into a compact feature vector, for example MFCC, PLP, Spectrogram, \emph{etc}. (ii) The next step is to learn an acoustic model, usually referred to as an encoder to associate these features to higher level speech or orthographic units (phonemes or graphemes). This is achieved either using Gaussian Mixture Models (GMM) \cite{khan18_sltu} or a Deep Neural Network (DNN) \cite{hannun2014deepspeech,amodei2016deepspeech2,li2019jasper} with HMMs, where the GMM/DNN acts as a function approximator and the HMM adds context capabilities to the encoder. With advances in deep learning, Long Short Term Memory networks (LSTMs) have replaced both the GMM/DNN and HMM models, since LSTMs act as function approximators and also learn context-dependence. The two most popular training frameworks used for acoustic modeling are Expectation-maximization (EM) \cite{dempster1977maximum(EM)} and Connectionist Temporal Classification (CTC) \cite{graves2006connectionist} respectively. Using the CTC loss function GMM/DNN-HMM system can be replaced by powerful auto-regressive models like LSTMs and Transformers.
(iii) Lastly, the output speech units are conditioned on a pre-trained language model (LM) \cite{heafield2011kenlm,BERT} such that the output sentence has a low perplexity. 

Traditional techniques optimize the ASR task in three steps independently as described above. Both GMM-based and early DNN-based techniques relied on the following language resources (i) A large corpus of speech transcribed by native speakers for training, (ii) a pronunciation lexicon mapping words in the language to phonemes, (iii) a phone set for the language,  (iv) a large corpus of text for the language model (v) transcribed data for evaluation. 

With the the availability of large annotated datasets for some languages, current state-of-the-art (SOTA) techniques solve the task in an end-to-end fashion \cite{gulati2020conformer,modernsynnaeve2019end,han2020contextnet} via supervised learning using only transcribed speech, in which sounds are directly mapped to graphemes/sub-word-units present in the language \cite{zhou2021acousticsubwordunits}. This forgoes the need for a lexicon and language model, which are learned implicitly by the model. Although end-to-end training seems attractive for low-resource languages where lexicons and text data may not be available, these models typically require larger amounts of labeled speech data than their traditional counterparts to reach the same level of performance \cite{googlestreamingmultilingual,toshniwal2018multilingual1}.

Recently, self supervised learning (SSL) has been used to pre-train an encoder on unlabelled data using contrastive loss. In this setup, the model learns a general high-level contextual representation of the input data which can potentially be used for any downstream task. The training process is divided into two steps: (i) learning an encoder which maps the raw audio to a high level compact representation, usually done using Convolutional Neural Networks (CNNs) (ii) Reconstruct the future frames given the high level features, using a strong auto-regressive model such as Transformers. Experiments by \cite{baevski2020wav2vec} have shown that quantizing the high level features before computing the contrastive loss leads to performance gains. These models have achieved SOTA performance in low resource scenarios and have been shown to perform on par, if not better, in high resource settings \cite{baevski2020wav2vec,hsu2021hubert,chung2021w2v}.



\section{Multilingual ASR models}

In this section, we survey recent research on multilingual ASR models. We categorize multilingual ASR models based on the data that they can be trained on - labeled data, in which each word that is spoken is carefully transcribed by a native speaker, and unlabeled data, which consists of speech without any transcriptions. Labeled data is expensive to create and may not always be feasible to collect if access to native speakers is challenging. Unlabeled data may be available 'in the wild' for many languages, in the form of user-generated content, such as on YouTube.

\subsection{Models that use only labeled data}

\subsubsection{Acoustic models}

\cite{heigold2013multilingualAM6} presents an extensive study comparing three major approaches towards building multilingual ASR using the GMM or DNN (or combined) and HMM setup. 

\begin{itemize}
\item Language independent feature extraction: In this approach, feature extraction is learned independent of the languages, so the feature extractor can be used for any language, after which an ASR system can be built on top of it. A major drawback of such a system is that it is not optimised jointly.

\item Transfer learning: In this approach, a model's learnable weights are initialised from a model trained on a high-resource language, followed by a separate fine-tuning step.

\item Multi-task learning: In this setting, one unified model is optimized/trained to perform $n$ tasks, without any separate fine-tuning step. In our case, $n$ is the number of different languages \emph{i.e.,} and each language is treated as a separate task.

\end{itemize}

\cite{heigold2013multilingualAM6} conduct experiments on 11 languages with a total duration of around 10000 hours for training and 35 hours for testing. The authors first compare, on all the 11 languages, GMM-HMM and DNN-HMM based monolingual systems and observe that the later performs better. Therefore they use DNN-HMM architecture for cross-lingual and multilingual experiments. (i) In the cross-lingual experiments, the DNN-HMM model is first trained on 3000 hours of English and then fine-tuned in two settings, either fine-tuning the top two layers or all layers for each language. Fine-tuning the top two layers avoids over-fitting for low resource languages. (ii) In the multilingual experiments, the DNN-HMM model is trained on combined data of all the 11 languages. It performs better than the cross-lingual setup consistently on all the languages. This study shows that \textbf{multilingual models trained by combining data of all languages performs better than fine-tuning an English model on the target languages.} 



\cite{thomas2010crossAM2} proposes using trained Multi-Layer Perceptron (MLP) features in cross-lingual settings \emph{i.e.,} adapting the learned MLP to a new language using phone mapping without re-initializing the last layer. The authors train an MLP on a combination of German and Spanish and after the phone mapping fine-tune the MLP on 1 hour English dataset, which they treat as the low-resource language. \cite{thomas2012multilingualAM1} proposes training a four-layered MLP for feature extraction on a German and Spanish followed by fine-tuning on English. In this setup, the first three layers are shared among different languages and the last layer is language specific. This results in in substantial (30\%) performance gain compared to the baseline monolingual GMM-HMM models. This work shows that \textbf{feature extraction and phone-mapping followed by fine-tuning of the target language results in significant gains over the monolingual model}, possibly because the source and target languages are from the same language family. 


\cite{tuske2013investigationAM4} compare systems built using Bottleneck (BN) features and MFCC features on datasets consisting of 150 hours of speech in German, Spanish, and English. They experiment with two setups (i) cross-lingual: fine-tuning the pre-trained MLP model from one language to a target language. They report a slight performance gain using BN features over monolingual baselines. (ii) multilingual: train one MLP model with all the languages combined. Their experiments show consistent improvement on all the three languages using BN features compared to MFCC features. 


\cite{tuske2013multilingualAM3} perform extensive experimentation to study the scalability of BN features using four languages \emph{i.e.,} French (317 hours), English (232 hours), German (142 hours), and Polish (110 hours) and train five and nine layer MLP models. The experiments show that BN features trained in multilingual settings perform better compared to BN features trained on the target language only. Lastly, jointly training a language dependent hidden layer with multilingual BN features performs best. This shows that the \textbf{choice of features may play an important role in cross-lingual transfer, and that training features on all languages performs better than training only on the target language.}


\cite{ghoshal2013multilingualAM5} trains the DNN directly, in the hybrid DNN-HMM setup, in sequential order and re-initialize the softmax layer for each language \emph{i.e.,} they train on one language then fine-tune on the second language, followed by fine-tuning on the third language and so on. They experiment with seven European languages in their setup with a total duration of 140hours for training and 14 hours for testing. Their setup shows improvement over the monolingual GMM-HMM and DNN-HMM baselines, however it is not clear whether the improvement is due to sequential training or the additional languages and training data added in each step. 

\cite{huang2013crossAM7} train a shared-hidden-layer multilingual DNN (SHL-MDNN), setup on four languages \emph{i.e,} French (138 hours), German (195 hours), Spanish (63 hours), and Italian (63 hours). Their experiments show that multilingual training improves over monolingual training on all languages. Furthermore they also train models on two unseen languages, English and Chinese, using the trained DNN as a feature extractor. Cross-lingual transfer performs better than training models for English and Chinese from scratch, which shows the learned features are transferable to unseen languages also. It is  interesting to note that \textbf{the transfer performs well for Chinese too, which is not in the same language family as the languages used to train the feature extractor}.

\cite{graves2013hybridLSTMbeatsDNN} shows that LSTM is a better choice of architecture than DNNs for ASR. Based on this, \cite{zhou17_interspeechAM8} study LSTM in the shared-hidden-layer multilingual LSTM (SHL-MLSTM) setting and compare it to SHL-MDNN. They show that monolingual SHL-MLSTM performs comparable to multilingual SHL-MDNN and multilingual SHL-MLSTM consistently outperforms multilingual SHL-MDNN on all the seven languages. SHL-MLSTM with residual learning helps them to train deeper networks, which results in even better performance. They hypothesize that residual learning helps overcome the gradient vanishing problem, indicating that \textbf{the choice of architecture is important while building larger, deeper models trained on multiple languages}. They also find that \textbf{the performance increases as the number of languages in the multilingual model increase}.

The Interspeech 2018 ASR challenge \cite{srivastava2018interspeech} compared models built for three Indian languages (Gujarati, Tamil and Telugu), in which baselines were monolingual models built using HMM-GMM, DNN-HMM and TDNN techniques. Most systems that improved over the baselines used multilingual training, that is combining data from all languages to build a single model and decoding using language specific Language Models \cite{billa2018isi}. \cite{sailor2020multilingualmtlsol} work on the same dataset and train four models (i) monolingual, (ii) multilingual, (iii) MTL with Language ID, and (iv) MTL with phoneme recognition. Their experiments show that using language ID as an auxiliary task degrades the performance of multilingual ASR system and propose that phoneme recognition is better choice. Adding \textbf{phoneme recognition as an auxiliary task gives small performance gains compared to the multilingual model}.









One other approach is to replace the GMM-HMM and train acoustic models (LSTMs) using the CTC loss function. \cite{tong2017multilingualctc1} studies training on multiple languages using the CTC loss function \emph{i.e.,} for English (81 hours), French (120 hours) and German (136 hours). In their experiments \textbf{training on all the languages combined performs worse compared to monolingual models, and fine-tuning multilingual models on target languages performs better than monolingual models}. They also observe performance gains when the trained multilingual model is fine-tuned on two low-resource unseen European languages. \cite{muller2017phonemicctc2} shows similar results that feeding language vectors during training performs better than training multilingual models by just combining data.

\subsubsection{seq2seq models}

\cite{watanabe2017language} used the hybrid attention/CTC framework to study multilingual training on 10 languages, with 1327 hours of speech for training and 150 hours for testing. The authors find that \textbf{the performance of some high resource languages degrades in multilingual settings} compared to the monolingual baselines, due to unbalanced training data. To overcome the imbalance, three techniques are proposed in the literature \emph{i.e.,} (i) Appending language ID vector in the input \cite{toshniwal2018multilingual1}, which performs better than the baseline, (ii) Sampling, and (iii) combining both, results in best performing models. So, \textbf{the degradation in performance for high-resource languages can be overcome by using sampling and language ID information}.

\cite{toshniwal2018multilingual1} studies multilingual training on 9 Indic languages, with around 1500 hours for training and 90 hours for testing. In their work, the authors experiment with four different models using Listen, Attend and Spell (LAS) \cite{ListenAttendandSpell} as the architecture. (i) monolingual (baseline) models for each language, (ii) A multilingual/joint model combining all the 9 languages, (iii.) A single model for ASR and predicting the language ID (iv.) A multilingual model conditioned on the language ID as an input. The language ID conditioned model gave the best performance, showing that \textbf{including language ID information helps training multilingual ASR systems}. They hypothesize that this is because the model reserves different parts of the network for different languages, while also being able to share language invariant features.


\cite{zhou2018multilingual3} shows similar findings using the transformer based architecture with experiments on six languages from the CALLHOME corpora. Their extensive experimentation with classical SOTA techniques shows that transformer based models perform better than \emph{i.e.,} Mono-DNN, Mono-LSTM, SHL-MDNN, SHL-MLSTM, and SHL-MLSTM-residual models. \cite{shetty2020improvingtransformermsrdata} also use transformer based models to show  performance gains when using language ID. They conduct experiments in three Indian languages \emph{i.e.,} Gujarati, Tamil, and Telugu. Lastly, they show that retraining the multilingual model on target languages further improves the performance compared to the universal multilingual model. 

\cite{googlestreamingmultilingual} learn an adapter module after each layer for each language separately for streaming E2E ASR. Similarly to \cite{toshniwal2018multilingual1}, they use nine Indic languages for training and testing. The authors train a system combining (i) multilingual RNN-T, (ii) language ID, and (iii) adapter modules. Both (ii) and (iii) performs best on low resource and on par in high resource languages. Adapter modules increase the overall model size compared to language ID training, however \textbf{there is a small performance boost for both low and high resource languages when an adapter module is learned for each language separately}.

The multilingual models surveyed so far are trained with fewer than 10 languages, in contrast to the massively multilingual models in NLP that are trained on 100+ languages. Recently \cite{pratap2020massivelymultilingual2} studies training of models of size up to 1 billion parameters on 51 languages, the largest till date. The total duration of speech data is around 16000 hours for training and 1000 hours for testing. The authors experiment with (i) a multilingual/joint model combining all the 51 languages, (ii.) a multilingual model conditioned on the language ID as an input and, (iii) multi-head model with 6-language-clusters ie. 6-different-decoders based on language family groupings. The multi-head model gives the best performance because \textbf{training the same decoder for multiple languages is beneficial if the languages are similar}.



\subsection{Models that use unlabeled data} \label{section:unsupervised}

Pre-training can be done on both labelled and unlabelled speech. Pre-training on unlabelled speech data is more practical for low-resource languages because it is easier and less expensive to find unlabeled speech data. 

\cite{conneau2020unsupervisedssl1} studied self-supervised multilingual training on 53 languages (56,000 hours) using the XLSR model. The authors train two variants of Wav2Vec2.0 \cite{baevski2020wav2vec} style models (base and large), and perform a series of experiments to study the effectiveness of multilingual SSL training. 
(i) Pre-training (pre1) (on English only), which outperforms monolingual models on low resource languages. 
(ii) Multilingual pre-training (pre2) on 10 languages (800 hours), with the same amount of data as in pre1, which outperforms pre1. This shows that \textbf{pre-training on multiple languages is a better choice rather than just adding more data from one language.} 
In both the cases, performance degrades on high resource languages. To address this degradation, they pretrain a larger model with 24 transformer blocks on 53 languages (56,000 hours) and this model shows improvements in high resource languages, in comparison to the pre2 model. This indicates that \textbf{SSL may need larger model capacity to show improvements for high resource languages}.
(iii) The Pre2 learnt representation transfers well to unseen low resource languages in low resource (50 hours)
(iv) Fine-tuning one model for all languages works a little worse on the base model and works slightly better on the large model compared to a monolingual model for each language. This again indicates that models require large capacity to show performance gains on larger datasets and a higher number of languages. (v) Finally, they show that \textbf{pre-trained models transfer well if the target language is similar to languages used for pre-training.} 

\cite{gupta2021clsrilssl2} extends these findings by training a model on Indic languages. They use around 10,000 hours for pre-training out of which the majority of the data is in Hindi (50\%). Their experiments results are similar to \cite{conneau2020unsupervisedssl1}, that \textbf{multilingual pre-training outperforms monolingual training}, even in the case of Hindi, which is a large dataset. 

\cite{javed2021towards} create a speech corpus of 17000 hours of unlabeled data in 40 Indian languages, including several very low resource languages. They use this data for pre-training multilingual ASR models and fine-tune the models on 9 Indian languages that have labeled data, leading to SOTA results on multiple benchmarks. Their findings suggest the following: (i) fine-tuning can be performed either monolingually, that is one-by-one for each target language, or combining all languages together. \textbf{Jointly fine-tuning a single model with all languages performs as well as multiple models fine-tuned separately}. 2. \textbf{The size of pre-training and fine-tuning datasets affect the accuracy of the final model} - contrary to previous findings (\cite{baevski2020wav2vec}), fine-tuning with just ten minutes of speech does not lead to high performance. 3. Building on \cite{conneau2020unsupervisedssl1}, \textbf{pre-training with a diverse set of languages improves performance on languages not present in the pre-training data}, which indicates that multilingual models may be useful for building speech technologies for very low resource languages, provided that some data for fine-tuning and evaluation is collected. Although these experiments do not include comparisons with monolingual models, the models beat previous SOTA models in terms of performance, which in turn are better performing than monolingual models \cite{srivastava2018interspeech}. 

\cite{wang2021unispeechpseudo3} proposes to combine supervised learning with labeled data and the CTC loss and SSL with the WAV2VEC2.0 loss in a Multitask Learning (MTL) scheme which they call UniSpeech. This helps the model to learn representations which are useful for speech recognition \emph{e.g.,} phoneme identities or invariance to background noise and accents. Furthermore they randomly replace the contextual representation with their quantized counterparts from the encoder when calculating the CTC loss. According to them this helps aligning the codewords from codebooks to ASR-specific features. The authors perform three sets of experiments based on one or multiple languages used during the pre-training and fine-tuning phase. (i) one-to-one (ii) many-to-one (iii) many-to-many. They report performance gains when using UniSpeech over XLSR \cite{conneau2020unsupervisedssl1} and CTC based model on all the three settings, showing that \textbf{the MTL setup is useful for cross-lingual transfer}. One downside of this learning scheme is that the learned features do not transfer to other tasks \emph{e.g.,} emotion or speaker recognition, as the features are specific to the ASR task. 


\cite{billa2021improving} compares ASR systems for three languages, Farsi, Kazakh and Lithuanian built using the following configurations: (i) in-domain labeled training data, which acts as the baseline (ii) mined data from Youtube in each language, followed by fine-tuning with in-domain data (iii) pooling data from 1 and 2 followed by fine-tuning (iv) transfer from a high-resource Arabic model by using the labeled data in each language. They find that fine-tuning either the model with in-domain or pooled data improves over the baseline, showing that \textbf{even unlabeled out-of-domain data can be used to improve performance over models built with only labeled in-domain data}. They also show that the fine-tuning approaches are comparable to transfer learning from a high-resource language.


Another technique for using unlabeled data is pseudo labeling, in which untranscribed data is transcribed using an existing ASR system. This is followed by a step to remove noisy labels, after which the transcribed data is added back to train the ASR model, leading to more data to train the model without investing resources for manual transcription. \cite{javed2021towards} shows that pseudo labelling for ASR \cite{kahn2020selfpseudo1} does not improve performance over 10 Indic languages compared to models built using only transcribed speech. They hypothesize that the generated pseudo labels are noisy due to the difference in distribution between the labeled and unlabeled speech, which is a common scenario in the real world. SSL may be a better technique for addressing domain and distribution differences in labeled and unlabeled data. Another drawback of the pseudolabel technique compared to SSL is that it is designed to be used within the same language and does not transfer.


\cite{zhang2021xlstpseudo2} addresses the lack of transfer across languages by using self-training to learn frame-level representations rather than generating the pseudo labels and call the algorithm Cross-lingual Self-training (XLST). XLST assumes that frame-level acoustic representations could be shared in some degree across different languages.  In their setup, instead of generating pseudo labels the authors generate frame embeddings from a pre-trained ASR model (target-network). Using this technique the authors show improvements over the XLSR model  \cite{conneau2020unsupervisedssl1} but the performance is worse compared to the UniSpeech model \cite{wang2021unispeechpseudo3}. We speculate that XLST performs worse than UniSpeech because UniSpeech optimizes the output representations directly on the ASR task using the CTC loss during the training. For similar reasons XLST performs better than XLSR because XLST use a pre-trained target network whose representations are optimized on the ASR task.





\section{Discussion}

\subsection{Multilingual vs. monolingual models for low-resource languages}

From the papers surveyed, it is clear that multilingual models perform better than monolingual counterparts trained with the same amount of data for a single language. Combining the data of all languages available during pre-training also improves performance compared to using multiple languages only during fine-tuning. 

\subsection{Techniques and architectures}

Fine-tuning plays an important role in the accuracy of multilingual models, with multiple studies showing that models fine-tuned to a target language perform better than simply combining data from all languages. Techniques that can be used for further improvements include phone-mapping, using a feature extractor trained on multiple languages and using a common decoder for related languages. Although proximity in terms of language family may play a role in accuracy, some studies show that unrelated languages also benefit from common feature extractors. Multitask Learning setups that use phoneme recognition or Language Identification as auxiliary tasks seem to improve performance only slightly over multilingual models trained jointly with all languages.

\subsection{Performance on high-resource languages}

When building a multilingual model that can be re-used across different languages, it is desirable that the performance on high-resource languages does not degrade while making improvements over low-resource languages. Some studies show that this degradation indeed occurs, however, there are strategies such as sampling and using language ID information that can be used to alleviate the problem. In the case of SSL, it seems that pre-training can help improve performance on high-resource languages as well.

\subsection{Factors that influence cross-lingual transfer}

Many factors affect the overall accuracy on target languages, including the choice of features and the languages that the features are trained on and the choice of architecture while building models trained on multiple languages. Some studies show that performance increases with the number of languages in the model, provided that the model is large and deep enough. In addition to training common feature extractors, it is also seen that training the same decoder for multiple languages performs better if the languages are similar.

\subsection{Role of SSL and unlabeled data}

Although research on this topic is nascent, it is clear that the choice of pre-training data matters. Pre-training from a diverse set of languages, or languages related to the target language is better than restricting it only to one language such as English even if the total pre-training data remains the same. Interestingly, pre-training with a diverse set of languages has been shown to improve performance on languages that are not present in the training data, though more experimentation is needed to study this further. It has also been shown that the pre-training data can be collected from a different domain than the target data and still improve performance. Finally, the size of pre-training and fine-tuning data matters - although it is possible to exploit unlabeled data for pre-training, the fine-tuning data needs to be of a reasonable size to get performance improvements. It is therefore important to invest in a minimum amount of labeled data for fine-tuning for target low-resource languages, unless there is a language in the multilingual model that is very closely related to the target language.

SSL and the use of unlabeled data are exciting directions for low-resource ASR, particularly when labeled and unlabeled data are from different domains. However, this approach shows some degradation for high-resource languages, that can potentially be addressed by increasing model capacity.

\subsection{Open problems and challenges}

Although cross-lingual transfer and SSL are promising directions for low-resource language ASR, there are some concerns that still need to be addressed by the research community. More research is need to determine whether these techniques can be used for very low-resource languages that have only a few hours of data, since labeled data is required for fine-tuning, transfer learning as well as evaluation. 

Studies on multilingual ASR have been limited to a few datasets and languages mainly from the Indo-European language family, and more research is needed to determine whether results hold true across languages. Evaluation benchmarks for multilingual ASR should be designed to cover diverse languages in terms of language family and dataset sizes (both labeled and unlabeled). Building such evaluation benchmarks is expensive but would be a crucial investment for improving low-resource language ASR.

Most multilingual models are trained on less than 10 languages and have not reached the size (in terms of languages) of their NLP counterparts that are trained on over 100 languages. An interesting direction of study would be the implications of larger models covering more languages on performance on high resource languages. As models become larger in terms of parameters, training time and training data required, it would be prudent to continue to compare performance with simpler monolingual models, particularly for low-resource languages.

SSL in NLP not only transfers across languages but also across tasks. Transferring representations learned by ASR models trained using SSL is an interesting future direction of research, in which some progress has been made already.

\section{Conclusion}

In this paper, we survey more than 40 papers on multilingual ASR, in which either multiple languages are trained in a single model, or cross-lingual transfer learning is used to improve performance on low-resource languages. Till recently, ASR models were only able to exploit transcribed data, however, with the advent of Self Supervised Learning, they are now also able to use unlabeled data which is an exciting development for low-resource languages. 
We address questions about whether multilingual models are indeed superior in performance to monolingual models for low-resource languages, as well as for high-resource languages. We distill key findings from research in multilingual ASR to describe factors that influence cross-lingual transfer and SSL, and provide recommendations for future research.

\section{References}
\bibliographystyle{lrec2022-bib}
\bibliography{lrec2022-example}


\end{document}